%% file: main_arxiv.tex
\documentclass[runningheads]{llncs}

\usepackage{graphicx}
\usepackage{multirow} 
\usepackage{geometry} 
\usepackage{threeparttable} 
\usepackage{graphicx}
\usepackage{booktabs} 
\usepackage{multirow} 
\usepackage{makecell} 
\usepackage[accsupp]{axessibility}  
\usepackage{xcolor} 
\usepackage[table]{xcolor}
\usepackage{relsize}
\usepackage[pagebackref,breaklinks,colorlinks]{hyperref}
\usepackage{orcidlink}
\usepackage{graphicx}
\usepackage{microtype}
\usepackage{algorithm, algorithmic}
\usepackage{pifont} 
\usepackage{subcaption}
\usepackage{cite}
\usepackage{bbding}

\input{macro}   

\begin{document}
	
	\title{\sysname: Pushing the Limit of Spatial Inteligence with Structured Scene Reasoning} 
	
	\titlerunning{\sysname: Pushing the Limit of Spatial Inteligence with Structured Scene Reasoning}
	
		
	\author{
		Yi Zhang\inst{1}\textsuperscript{*} \and
		Youya Xia\inst{1}\textsuperscript{*} \and
		Yong Wang\inst{1} \and
		Meng Song\inst{1} \and
		Xin Wu\inst{2} \and
		Wenjun Wan\inst{2} \and \\
		Bingbing Liu\inst{1} \and
		Aixue Ye\inst{1}\textsuperscript{\Envelope} \and
		Hongbo Zhang\inst{1} \and
		Feng Wen\inst{1}
		}
	
	\authorrunning{Y. Zhang et al.}
	
	\institute{
		Foundation Model Department, Huawei \and
		Central Media Technology Institute, Huawei \\
    	\textsuperscript{*}Equal contribution. \quad \textsuperscript{\Envelope}Corresponding author. \\
		\email{\{zhangyi432,xiayouya,wangyong279,songmeng6,wuxin79,wanwenjun3,\\
		liu.bingbing,yeaixue,zhanghongbo888,feng.wen\}@huawei.com}
		}
	
	\maketitle
	
	\begin{abstract}
		While Multimodal Large Language Models (MLLMs) excel in semantic tasks, they frequently lack the "spatial sense" essential for sophisticated geometric reasoning.
		Current models typically suffer from exorbitant modality-alignment costs and deficiency in fine-grained structural modeling precision.
		We introduce \sysname, a framework designed for \textbf{S}tructured \textbf{S}cene \textbf{R}easoning that seamlessly integrates 2D and 3D representations via a lightweight alignment mechanism.
		To minimize training overhead, our framework anchors 3D geometric features to the large language model's pre-aligned 2D visual semantics through cross-modal addition and token interleaving, effectively obviating the necessity for large-scale alignment pre-training.
		To underpin complex spatial reasoning, we propose a novel scene graph generation pipeline that represents global layouts as a chain of independent local triplets defined by relative coordinates. This is complemented by an incremental generation algorithm, enabling the model to construct "language-model-friendly" structural scaffolds for complex environments.
		Furthermore, we extend these capabilities to global-scale 3D global grounding task, achieving absolute metric precision across heterogeneous data sources.
		At a 7B parameter scale, \sysname~achieves state-of-the-art performance on multiple spatial intelligence benchmarks, notably scoring 73.9 on VSI-Bench. Our approach significantly outperforms much larger models, demonstrating that efficient feature alignment and structured scene reasoning are the cornerstones of authentic spatial intelligence.
		
	\end{abstract}
	
	\section{Introduction}
	\label{sec:intro}
	
	Humans possess an innate "spatial sense", a cognitive faculty that allows us to implicitly reconstruct 3D environments, estimate metric distances, and predict temporal-spatial evolutions from simple 2D retinal observations. This ability is not merely about recognizing objects but about building a consistent mental scaffold of the physical world. While MLLMs have achieved remarkable success in general visual understanding and open-ended dialogue, they still fundamentally struggle with tasks requiring precise geometric reasoning. As noted in recent evaluations, even state-of-the-art models often fail at basic spatial tasks like distance estimation or maintaining layout consistency across multiple viewpoints.
	
	The limitations of current spatial intelligence in MLLMs stem from two primary challenges. First, most existing models attempt to incorporate external spatial representations (such as 3D point clouds or depth maps) through heavy pre-training and alignment stages. This paradigm necessitates large-scale, modality-specific data to bridge the gap between geometric features and language embeddings, incurring significant computational costs. There is a pressing need for a new spatial feature alignment strategy that relieves this burden by leveraging the inherent alignment already established between 2D visual information and the Large Language Model (LLM). Second, existing models are typically trained on general spatial reasoning Question-Answering (QA) pairs, which focus on scene-level descriptions or quantitative questions but lack fine-grained, structured scene representations. However, building a structured internal model of a scene is a crucial prerequisite for complex reasoning. Much like humans naturally construct a mental scaffold of their surroundings before addressing spatial queries, an intelligent system must first master structured scene representation to achieve robust spatial cognition.
	
	\begin{figure}
		\centering
		\includegraphics[width=0.65\linewidth]{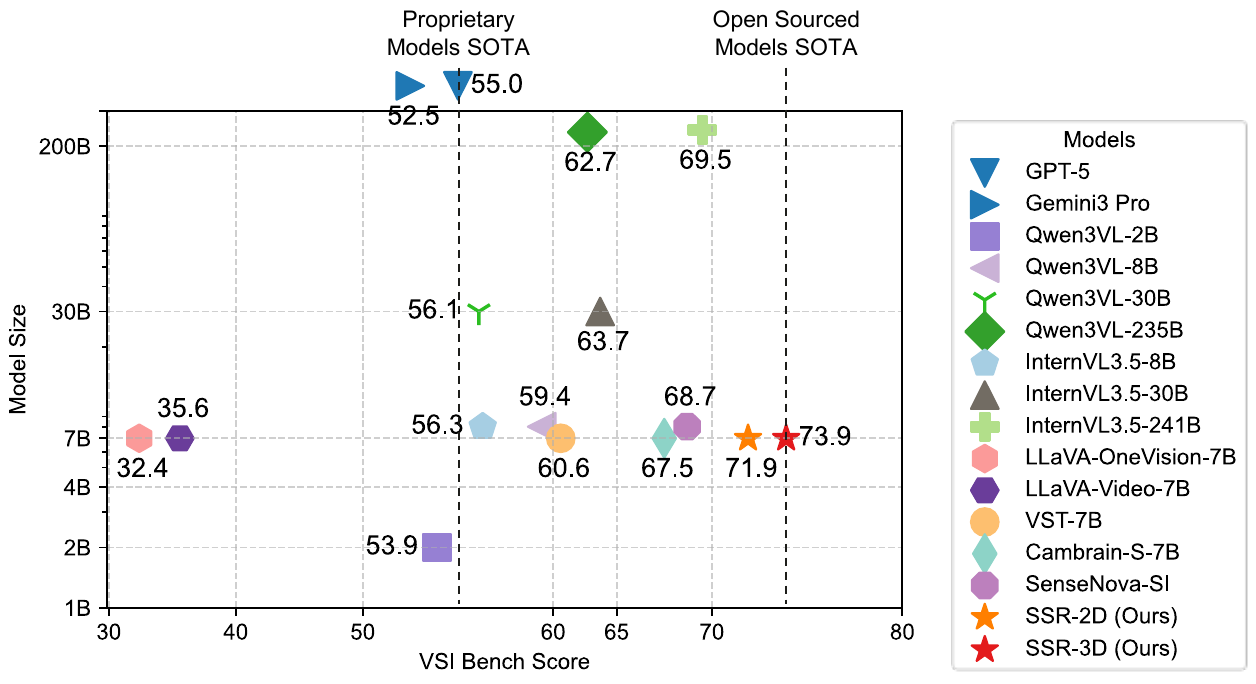}
		\caption{Comparison of model performance on VSI-Bench. \sysname~achieves the highest accuracy among all proprietary and open-source competitors. Notably, our 7B model outperforms significantly larger models, demonstrating superior parameter efficiency in spatial reasoning.}
		\label{fig:teaser}
	\end{figure}
	
	To tackle the first challenge, we propose a novel framework that incorporates both 2D and 3D scene representations into the LLM through a simple yet effective modality alignment mechanism. Specifically, we leverage the 2D vision features—which are already well-aligned with the LLM—to facilitate 3D alignment via a proposed two-stage strategy. In the first stage, we merge 2D features into the 3D spatial branch, effectively making the 3D geometric features "readable" by the LLM by anchoring them to known visual semantics. In the second stage, we introduce an interleaved token insertion method that alternates 2D visual and 3D spatial features on a frame-by-frame basis. This ensures that corresponding features from the same temporal instance are aligned within the LLM's token space, promoting fine-grained cross-modal interaction without the need for exhaustive, from-scratch modality alignment training.
	
	To resolve the second challenge, our motivation is to train the model to generate structured scene representations based on visual input, thereby acquiring fundamental spatial modeling abilities. Specifically, we train the model to generate \textbf{LocalCogMap} (\textbf{Local} \textbf{C}ognitive \textbf{M}ap), a carefully designed scene graph representation that discretizes local triplets into a $10 \times 10$ grid. By representing object spatial arrangements through relative and normalized coordinates, we translate abstract geometry into a discrete format.  By employing an incremental generation mechanism, we provide the LLM with a "language-model-friendly" structural scaffold. This allows the model to decompose complex global scenes into consistent local coordinates, mirroring the human process of building mental scene structures as a cognitive foundation for high-level spatial deduction. To complement these local structures with fine-grained metric precision, we also incorporate a 3D global grounding task into the fine-tuning stage. This enables the model to output object 3D bounding boxes at a global scale, effectively bridging the gap between symbolic relative arrangements and absolute metric grounding. Our contributions can be summarized as follows:
	
	\begin{itemize}
		\item \textbf{Efficient 3D-Aware Architecture:} We introduce a dual-branch MLLM architecture that integrates both 2D appearance and 3D geometric features. By leveraging inherent visual priors and a novel interleaved token insertion strategy, our framework achieves effective multi-modal alignment with significantly reduced training effort;
		\item \textbf{Structured Mental Modeling Paradigm:} We propose a novel spatial reasoning paradigm that integrates a localized scene graph formulation, termed LocalCogMap, with global 3D grounding. By supervising the model to directly generate structured representations and 3D object coordinates from visual inputs, we enable the construction of fine-grained 'mental scene graphs.' These representations serve as a robust cognitive foundation, significantly enhancing the model's capacity for complex spatial reasoning;
		\item \textbf{High-Quality Data and Open-Source Models:} We curate a large-scale structured scene representation dataset comprising approximately 190K samples, designed to bridge the gap between 2D perception and 3D geometric reasoning. In addition, we provide our pre-trained, high-efficiency spatial intelligence models to the community. By making these resources publicly available, we aim to establish a robust foundation for future research and push the frontiers of spatial reasoning in multimodal systems;
		\item \textbf{State-of-the-Art Performance:} \sysname~surpasses the performance of significantly larger models across diverse spatial reasoning benchmarks—most notably VSI-Bench~\cite{yang2025thinking} (Fig.~\ref{fig:teaser})—validating the superior effectiveness and architectural efficiency of our structural intelligence approach.
	\end{itemize}
	
	\section{Related Work}
	\label{sec:related}
	
	\subsection{Multimodal Foundation Models}
	Recent Multimodal Large Language Models typically employ a pre-trained vision encoder and an MLP-based projector to map visual embeddings into the language space~\cite{Qwen3-VL, zhu2025internvl3}. To push the performance ceiling, strategies such as dynamic resolution, high-quality data synthesis, and post-training via reinforcement learning—specifically GRPO~\cite{shao2024Deepseekmath} inspired by DeepSeek-R1—have significantly enhanced 2D reasoning capabilities. However, these models remain fundamentally 2D-centric, processing visual inputs as planar patches without intrinsic 3D structural representations. This limitation leads to substantial deficiency in spatial intelligence and 3D-aware reasoning tasks~\cite{yang2025thinking, cai2025holistic}.
	
	\subsection{Spatial Intelligence Foundation Models}
	To endow MLLMs with a deeper understanding of the physical world, recent research~\cite{cai2025scaling, yang2025visual, feng2025vica, fan2025vlm3r, yang2025cambrians, chen2025reasoning, jia2025omnispatial, wu2025spatialmllm, ma2025spatialllm, cheng2024spatialrgpt, chen2024spatialvlm, li2025sti, yang2025thinking} has shifted toward spatial intelligence and some try to propose noval architecture by integrating 3D-aware priors. These models fall into several architectural paradigms. Geometry-aware unified frameworks like VLM-3R~\cite{fan2025vlm3r} extract implicit 3D structure from monocular video and align it with language through over 200K reconstructive QA pairs. In contrast, Spatial-MLLM~\cite{wu2025spatialmllm} utilizes a dual-encoder architecture—combining a 2D visual encoder with a geometry-aware spatial encoder—and employs space-aware frame sampling to maximize scene coverage. Further, SpaceR~\cite{ouyang2025spacer} incorporates a map imagination module and utilizes Spatially-Guided RLVR to achieve strong performance on benchmarks like VSI-Bench. Despite these advances, a major bottleneck remains: existing state-of-the-art models typically rely on heavy spatial-alignment training, requiring massive reconstructive datasets or computationally expensive reinforcement learning to bridge the gap between language and 3D geometry. This leads to prohibitive computational and data labeling costs. In contrast, our work introduces a light-weighted spatial alignment architecture. By efficiently mapping spatial features to the MLLM's latent space without exhaustive reconstructive supervision, we maintain high spatial reasoning performance while significantly reducing the training overhead.
	
	\subsection{Structural Spatial Representations and Grounding}
	he efficacy of spatial intelligence is fundamentally anchored in scene representation and grounding consistency. Traditional 3D Scene Graph (SGs) reprsentations~\cite{mitra2024compositional,rosinol20203d,zemskova20253dgraphllm,wu20243d,xu2025geonav,chandhok2024scenegpt,chen2025schema,yin2024sg} often suffer from coarse granularity, failing to capture the fine-grained spatial orientations required for complex reasoning. Furthermore, 3D grounding in dynamic videos is hindered by the lack of stable reference frames, as current approaches~\cite{bai2025qwen3vltechnicalreport, guo2025seed15vltechnicalreport} typically anchor coordinates to single frames, leading to significant instability under camera motion. To address these limitations, we propose LocalCogMap, a "language-model-friendly" cognitive scaffold designed to discretize local spatial arrangements into structured, manageable representations. By further incorporating a 3D Global Grounding task, our approach effectively bridges the gap between symbolic relative reasoning and absolute metric precision, enabling robust spatial intelligence across long-horizon dynamic scenes.
	
	\section{Methods}
	\label{sec:method}
	
	\subsection{Model Architectrure}	
	\label{sec:model_arch}
	\subsubsection{Overview.} 
\begin{figure}
		\centering
		\includegraphics[width=0.95\linewidth]{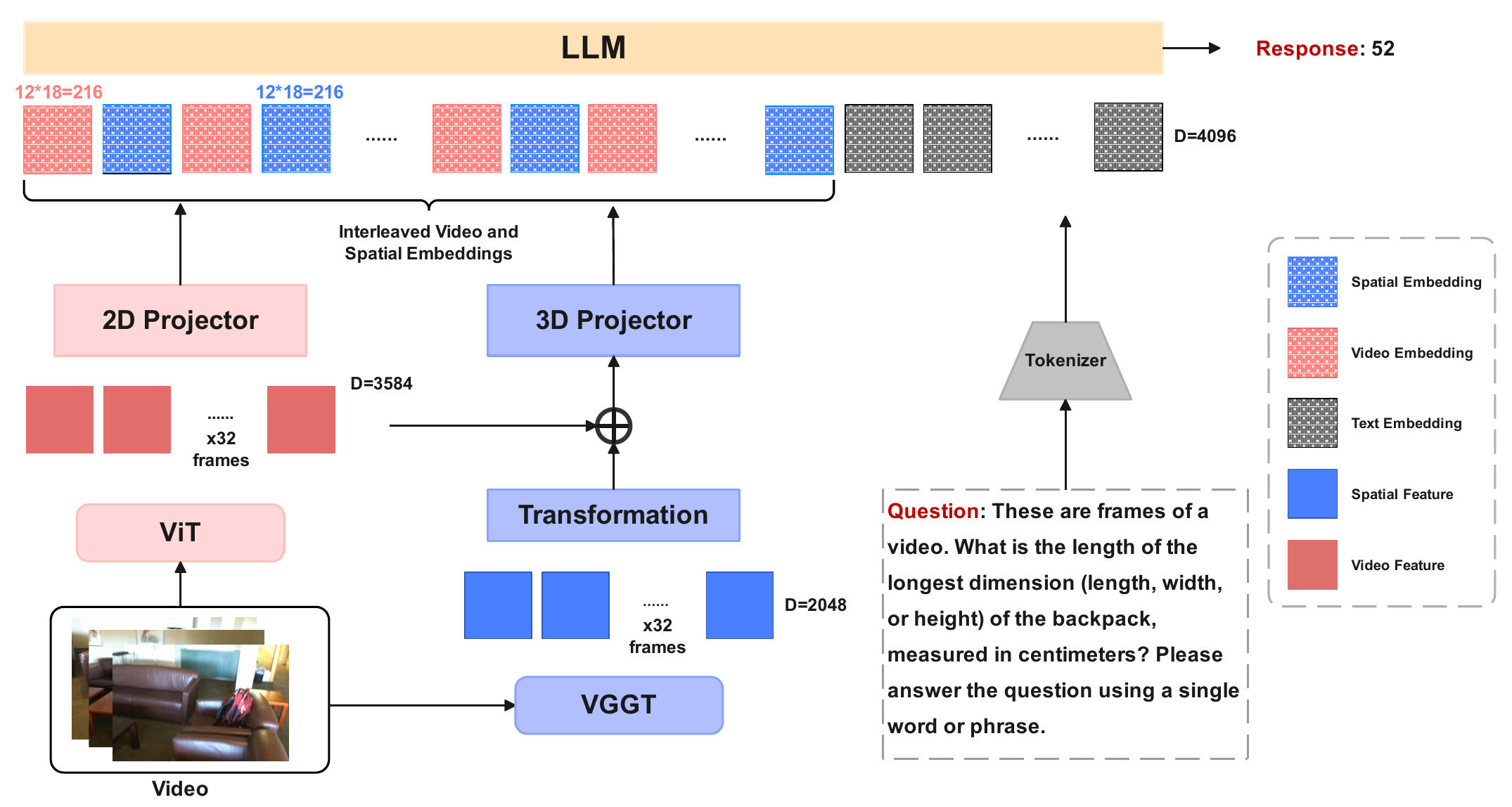}
		\caption{Architecture of \sysname-3D: It adopts a dual-branch architecture to jointly leverage 2D visual and 3D spatial cues. The 3D branch encodes geometric scene structure through dedicated spatial tokens, while the 2D branch processes image-derived visual features extracted by a vision encoder. Tokens from both branches are then interleaved and fused as input to the LLM, enabling unified multimodal reasoning over appearance and geometry.}
		\label{fig:29}
	\end{figure}

Fig.~\ref{fig:29} illustrates the overall architecture of \sysname-3D, our comprehensive dual-branch framework (a streamlined version, \sysname-2D, is detailed in Sec.~\ref{sec:train_strategy}). We propose a Multimodal Large Language Model (MLLM) architecture that seamlessly integrates 2D appearance features with 3D geometric cues. Crucially, by leveraging inherent visual priors and a novel interleaved token insertion strategy—in which visual and spatial embeddings from the same video frame are placed in adjacent positions—our framework achieves effective multimodal alignment with significantly reduced training effort. The architecture consists of two parallel branches: a 2D branch that processes appearance-based visual inputs extracted from video frames, and a 3D branch that fuses spatial and visual cues to produce structured spatial embeddings. These aligned vision and spatial representations are then seamlessly integrated with the language token embeddings of LLM and jointly fed into its decoder to generate the final output.

	\subsubsection{Spatial Feature Extraction.}
	To extract spatial features $\mathbf{s}$, we employ VGGT~\cite{wang2025vggt} as the core backbone for spatial encoding. Specifically, we uniformly sample $N=32$ frames $\{I_1, \dots, I_N\}$ from each training video sequence. Rather than utilizing the final semantic layers, we extract intermediate representations from the intermediate layer of VGGT encoder. This choice is motivated by the empirical observation that these mid-level features exhibit superior multi-view geometric consistency and spatial fidelity—properties that are indispensable for robust 3D scene understanding. The resulting spatial representation is formulated as:
	\begin{equation}
		\mathbf{s} = \Phi_{\text{vggt}}(\{I_1, \dots, I_N\}) \,,
	\end{equation}
	where $\Phi_{\text{vggt}}(\cdot)$ denotes the geometry-aware feature mapping derived from the attention blocks of the 23rd layer of VGGT encoder.
	\subsubsection{3D Feature Fusion.} Since the 3D spatial branch lacks large-scale pretraining, directly incorporating VGGT-derived~\cite{wang2025vggt} spatial features into this branch results in a significant representation gap relative to the visual features extracted from the pretrained 2D branch. To mitigate this misalignment, we introduce a lightweight transformation layer $\mathrm{MLP_{trans}}$ within the spatial branch that maps the spatial features $\mathbf{s}$ into the same embedding space as the visual features:
	\begin{equation}
		\mathbf{s}' = \mathrm{MLP_{trans}}(\mathbf{s}) \,.
	\end{equation}
	The transformed spatial embeddings $\mathbf{s}'$ are then fused with their corresponding visual counterparts ${\mathrm{ViT}}(\mathbf{v})$ encoded by ViT from the input videos $\mathbf{v}$ through element-wise addition before being injected into the input embedding space of the large language LLM:
	\begin{equation}
		\mathbf{s}^{\mathrm{fused}} = {\mathrm{ViT}}(\mathbf{v}) + \mathbf{s}' \,.
	\end{equation}
	This fusion strategy ensures that the resulting tokens retain complementary geometric structure and visual appearance cues, thereby implicitly establishing a cross-modal alignment pathway between 3D spatial representations and 2D visual semantics.
	\subsubsection{3D Spatial Branch.} Following the fusion of spatial and visual features, the resulting representation $\mathbf{s}^{\mathrm{fused}}$ is projected into the language embedding space via a dedicated 3D projector—structured analogously to the vision projector—using a lightweight $\mathrm{MLP_{3D}}$:
	\begin{equation}
		\mathbf{s}^{\mathrm{proj}} = \mathrm{MLP_{3D}}(\mathbf{s}^{\mathrm{fused}}) \,.
	\end{equation}
	\subsubsection{Multimodal Interleaved Insertion.} In contrast to conventional token insertion strategies—which typically concatenate modality-specific embeddings sequentially (\textit{e.g.}, all visual tokens followed by all spatial tokens)—we argue that such rigid sequential ordering impedes effective cross-modal alignment. In \sysname, all input tokens are uniformly indexed using Multimodal Rotary Position Embedding (M-RoPE). Specifically, given $T$ sampled video frames, both visual and spatial features are assigned sequential positions within the unified range $[0, 2T]$. Under the naive concatenation scheme, embeddings corresponding to the same temporal frame are separated by a fixed offset of $T$ in their positional indices. This large positional discrepancy introduces a strong inductive bias that disrupts fine-grained cross-modal interaction, particularly in the absence of an explicit alignment training stage between visual and spatial representations.
	
	To mitigate the misalignment caused by the introduction of an additional modality in the absence of large-scale pretraining, we propose a novel token insertion strategy that interleaves visual and spatial embeddings at the frame level. As illustrated in Fig.~\ref{fig:29}, for each video frame $t$, the corresponding visual embedding is immediately followed by its associated spatial embedding. This interleaved arrangement ensures that cross-modal representations originating from the same temporal instance share adjacent positions in the input sequence, thereby promoting fine-grained alignment without requiring explicit correspondence learning.
	
	\subsection{Scene Graph}
	
	\begin{figure}[t]
		\centering
		\begin{subfigure}{0.36\textwidth}
			\centering
			\includegraphics[width=\linewidth]{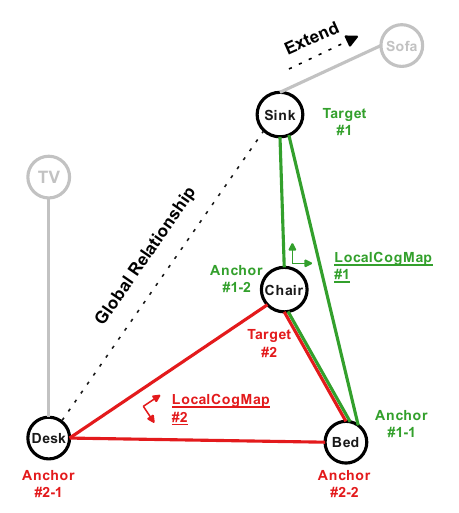}
		\end{subfigure}
		\hfill
		\begin{subfigure}{0.36\textwidth}
			\centering
			\includegraphics[width=\linewidth]{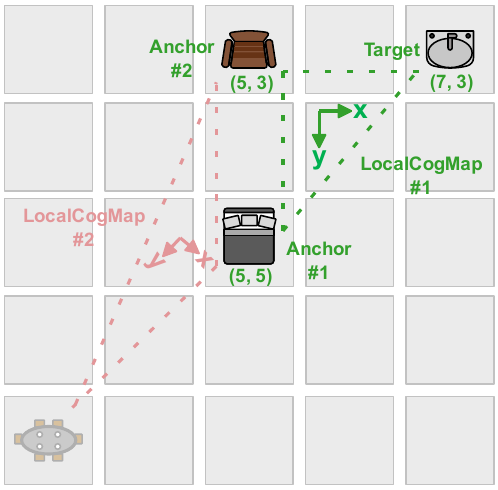}
		\end{subfigure}
		\caption{Global scene graph representation via LocalCogMap. \textbf{Left}: Global Scene Graph: Our proposed framework maintains global connectivity while redefining triplets as localized spatial units. \textbf{Right}: LocalCogMap Construction: Each triplet is modeled within a $10 \times 10$ grid established by two anchors. The target object is then normalized within this frame. This formulation ensures geometric consistency across the entire scene graph.}
		\label{fig:sg_localcogmap}
	\end{figure}
	
	\subsubsection{Scene Graph Formulation.}
	To enhance the spatial reasoning capabilities of MLLMs, a promising strategy is to cultivate the capacity for spatial mental modeling based on 2D inputs. Specifically, we aim to pre-train the model’s ability to generate scene representations as a prerequisite to complex spatial reasoning. This approach is motivated by a famous dictum: \textit{"What I cannot create, I do not understand."} While traditional methods utilize dense representations like depth maps or point clouds, these formats are often architecturally incompatible with LLMs, which are optimized for discrete, tokenized outputs and require extensive alignment to bridge the modality gap. Consequently, a primary challenge lies in formulating a scene representation that is "language-model-friendly"—meaning it must be both discrete and self-contained. To this end, we propose a novel scene graph structure that translates spatial configurations into a format that can be easily generated by LLMs.
	
	Existing scene graph representations generally fall into two categories. The first~\cite{mitra2024compositional, zemskova20253dgraphllm, wu20243d} employs \textbf{relationship-based graphs}, where edges are defined by qualitative spatial prepositions (\textit{e.g.}, "left of," "inside"). These models struggle to capture fine-grained spatial metrics, such as precise relative distances. The second category~\cite{rosinol20203d} utilizes \textbf{hierarchical graphs}, which organize concepts across varying levels of granularity. While effective for embodied AI applications, these structures still lack the precision required to represent detailed scene layouts. In contrast, as illustrated in the left part of Fig.~\ref{fig:sg_localcogmap}, our proposed structure maintains a graph-based framework for global connectivity but redefines the underlying triplets. Unlike traditional methods, each triplet in our graph is modeled within a localized coordinate system, which we term the LocalCogMap. As shown in the right part of Fig.~\ref{fig:sg_localcogmap}, the LocalCogMap utilizes a $10 \times 10$ grid established by two "anchor" objects. The location of a third "target" object is then normalized within this grid. For example, by positioning two anchors at fixed coordinates—such as $[5, 5]$ and $[5, 3]$—the target's relative position (\textit{e.g.}, $[7, 3]$) can be mapped precisely, ensuring spatial consistency. Currently, our formulation focuses on the bird’s-eye-view to cover the majority of reasoning tasks, though it can be extended to 3D environments.
	
	Distinguishing itself from conventional scene graphs that rely on pairwise relationships, the LocalCogMap adopts a \textbf{triplet-based system} that represents layouts quantitatively. By discretizing the coordinate system into a $10 \times 10$ grid, we significantly reduce the generative complexity for the LLM. Crucially, these local triplets are designed to overlap. This overlapping structure allows the model to generalize local spatial relationships into a coherent global representation, providing a robust framework for both local precision and global scene understanding.
	
	\subsubsection{Scene Graph Generation.}
	
	Given our proposed formulation, the challenge lies in instantiating LocalCogMaps for local triplets while maintaining global geometric consistency. To ensure that local relationships scale to the global context, we must carefully select which triplets to model. An exhaustive search—traversing all possible combinations of three objects—would incur a computational complexity of $O(N^3)$ with $N$ being the number of objects, making it intractable for LLM. Stochastic triplet sampling often leads to structural failures. In some instances, the scene graph partitions into disconnected components because no single triplet bridges separate object clusters. In other cases, the graph remains under-constrained: while nominally connected, the relative orientation between clusters remains undetermined. This lack of geometric rigidity precludes the unique deduction of coordinates between independent groups of objects. A detailed visualization of the two corner cases can be found in the Appendix S2.
	
	To mitigate these issues and construct a globally consistent representation, we propose an \textbf{Incremental Scene Graph Generation} algorithm. The core principle is to initialize the graph with a single triplet and incrementally incorporate remaining objects. At each step, the algorithm ensures that the location of a newly added object can be deterministically inferred from at least two existing anchors within the graph. The implementation details are provided in Alg.\ref{algo:isgg}.
	
	\begin{algorithm}[h]
		\caption{Incremental Scene Graph Generation}
		\label{algo:isgg}
		\begin{algorithmic}[1]
			\REQUIRE Set of 3D bboxes $O$, Threshold $\delta$
			\ENSURE Set of LocalCogMaps $\mathcal{L}$

			\STATE $\mathcal{L} \leftarrow \emptyset, \mathcal{V}_{in} \leftarrow \emptyset, \mathcal{V}_{out} \leftarrow O$

			
			\FOR{each triplet $\{o_i, o_j, o_k\} \subseteq O$}
			\IF{$\max(dist(o_i, o_j), dist(o_j, o_k), dist(o_i, o_k)) \le \delta$}
			\STATE $LCM_{init} \leftarrow \text{CreateLocalCogMap}(o_i, o_j, o_k)$
			\STATE $\mathcal{L} \leftarrow \mathcal{L} \cup \{LCM_{init}\}$
			\STATE $\mathcal{V}_{in} \leftarrow \{o_i, o_j, o_k\}, \mathcal{V}_{out} \leftarrow O \setminus \mathcal{V}_{in}$
			\STATE \textbf{break}
			\ENDIF
			\ENDFOR
			\WHILE{$\mathcal{V}_{out} \neq \emptyset$}
			\STATE Pick $u \in \mathcal{V}_{out}$
			\STATE $\{v_1, v_2\} \leftarrow \arg\min_{\{v_a, v_b\} \subseteq \mathcal{V}_{in}} (dist(u, v_a) + dist(u, v_b))$
			\STATE $LCM_{new} \leftarrow \text{CreateLocalCogMap}(u, v_1, v_2)$
			\STATE $\mathcal{L} \leftarrow \mathcal{L} \cup \{LCM_{new}\}$
			\STATE $\mathcal{V}_{in} \leftarrow \mathcal{V}_{in} \cup \{u\}, \mathcal{V}_{out} \leftarrow \mathcal{V}_{out} \setminus \{u\}$
			\ENDWHILE
			
			\RETURN $\mathcal{L}$
		\end{algorithmic}
	\end{algorithm}

\begin{figure}[h]
		\centering
		\includegraphics[width=\textwidth]{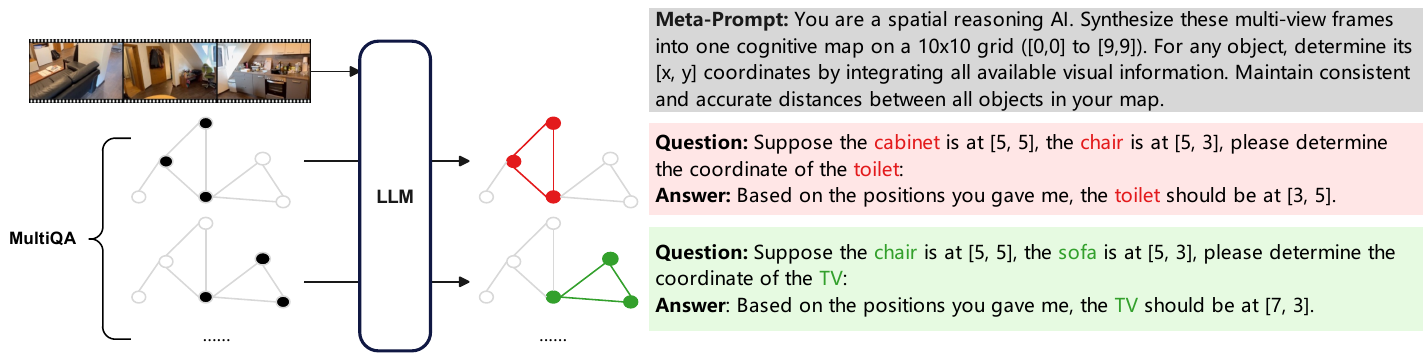}
		\caption{MultiQA-based scene graph generation. We transform global scene graphs into independent triplets. For each triplet, the LLM infers target coordinates relative to two anchors within a structured system context. Compared to dense captions, this decoupled QA format ensures scalability to complex scenes and reduces computational redundancy.}
		\label{fig:multiqa}
	\end{figure}

	Unlike traditional methods that treat scene graphs as abstract data structures, we focus on generating these graphs using LLMs. This requires converting the graph into a textual format compatible with the \textbf{next-token prediction} paradigm. We introduce a \textbf{MultiQA} pipeline, which decomposes the global scene graph into independent triplets. For each triplet, we construct a QA pair. Each sample prompts the model with a system context—defining the cognitive map and the generation task—and asks it to infer the coordinates of a "target" object given the known positions of two "anchors." An example of this MultiQA format is shown in Fig.~\ref{fig:multiqa}. Our decision to use decoupled triplets rather than a single, dense caption is twofold. First, scenes comprising dozens of objects will yield a LocalCogMap description that exceeds the effective context window of most large language models (LLMs). Second, a multitude of downstream tasks only demand a subset of the scene’s spatial data, making the generation of the complete graph for each query computationally redundant. Second, many downstream tasks only require a subset of the scene’s spatial data; generating the entire graph for every query is computationally redundant. Finally, this decoupled structure naturally supports Chain-of-Thought reasoning, as individual triplets can serve as intermediate "scaffolds" for more complex spatial deductions.

	\subsection{3D Global Grounding}
	
	The heterogeneity of coordinate definitions (\textit{e.g.}, origin location and axial alignment) across contemporary 3D global grounding datasets poses a substantial barrier to large-scale data curation. To mitigate this issue, we propose a unified 3D coordinate framework designed to provide consistent spatial representations for robust 3D global grounding.
	
	\subsubsection{Coordinate Definition.} We define a 7-DoF (Degree of Freedom) representation for target objects to balance descriptive precision with computational efficiency, parameterizing each object as a 7-tuple $\mathbf{b} = (x_c, y_c, z_c, l, w, h, \theta_{\text{yaw}})$. In this formulation, $(x_c, y_c, z_c)$ denotes the spatial center of the 3D bounding box within the global coordinate frame, while $(l, w, h)$ captures the geometric dimensions along the $X$, $Y$, and $Z$ axes to characterize physical scale. The yaw angle $\theta_{\text{yaw}}$ represents the angular displacement around the vertical $Z$-axis in a right-handed Cartesian system, with rotations following the right-hand rule and parameterized in radians to ensure numerical stability during model optimization. Notably, we omit roll and pitch angles as this 7-DoF representation provides a sufficiently unambiguous description for the vast majority of indoor and outdoor grounding scenarios while significantly reducing the complexity of the optimization space.
	
	
	\subsubsection{3D Global Grounding Coordinate Generation.}
\begin{figure}
		\centering
		\includegraphics[width=0.8\linewidth]{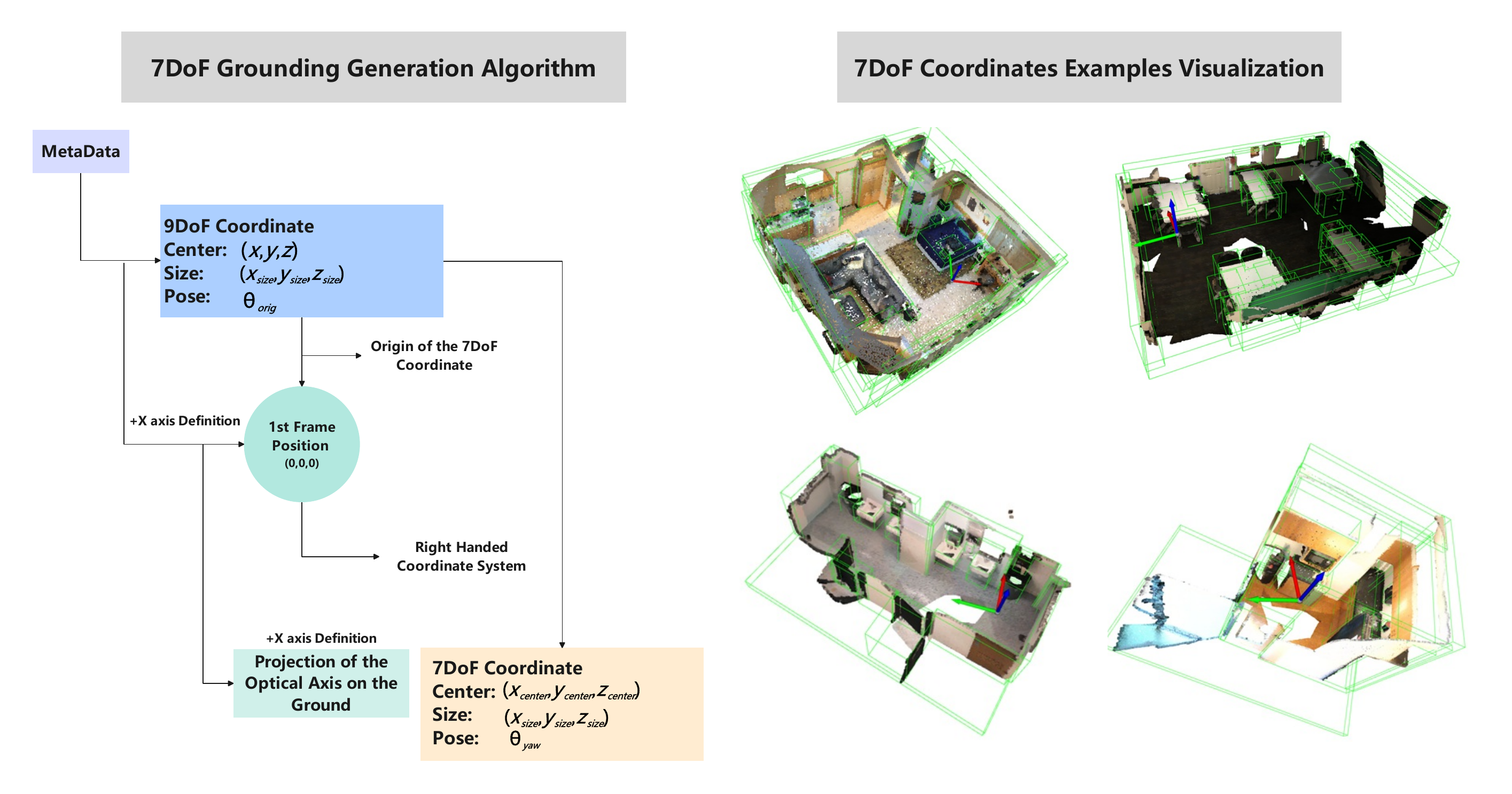}
		\caption{The visualization of 7-DoF coordinates generation algorithm. As illustrated in the generation pipeline, we define the origin of the global coordinate system as the camera position in the first frame, and align the positive direction of the X-axis with the projection of the optical axis onto the ground plane. The visualizations of the 7-DoF grounding results demonstrate that our proposed coordinate definition is both geometrically clear and highly adaptable across diverse scenarios and datasets.}
		\label{fig:3}
	\end{figure}

	Building upon the defined 7-DoF representation, this section elaborates on the transformation pipeline designed to map raw coordinates into our standardized coordinate system. As illustrated in Fig.~\ref{fig:3}, this normalization procedure is structured into a systematic three-step process. First, for object dimension ($x_{\text{size}}, y_{\text{size}}, z_{\text{size}}$), we derive scale parameters directly from the original 9-DoF poses provided in the metadata, as these dimensions remain invariant to coordinate system transformations. Second, to establish a consistent coordinate system origin ($x_{\text{center}}, y_{\text{center}}, z_{\text{center}}$) despite camera ego-motion and the lack of universal landmarks, we fix the origin at the optical center of the camera at the initial frame of the video sequence. Finally, for axis alignment and formalization, we define the positive $x$-axis as the projection of the camera's optical axis onto the ground plane at the first frame. This orientation simplifies pose transformations and projection matrices while reducing coordinate conversion complexity. The complete 3D coordinate frame is then formalized as a right-handed system based on this predefined origin and $x$-axis direction.
	Based on the proposed algorithm to unify the 3D global grounding coordinatge system, we can process 3D metadata such as ScanNet~\cite{dai2017scannet}, ScanNet++~\cite{yeshwanthliu2023scannetpp} or Arkitscenes~\cite{dehghan2021arkitscenes} to obtain large-scale 3D global grounding QA dataset.

	\subsubsection{Grounding Data Curation.}
	Given that 3D environments frequently contain multiple instances of the same semantic category, accurately localizing the specific target of interest is paramount. To address this challenge, we propose three distinct strategies for unambiguous object referral:
	
	\begin{itemize}
		\item \textbf{Proximity-based Reference:} Utilizing a specific environmental landmark as an anchor, we distinguish target objects based on their relative distance, such as identifying the instance nearest to or furthest from the reference anchor.
		
		\item \textbf{Direction-based Reference:} By establishing a reference position via an anchor object and a reference orientation via another, we localize targets by their relative angular placement, effectively querying objects located in a specific relative direction.
		
		\item \textbf{Temporal Appearance Order:} By leveraging the chronological order of an object's first appearance in the video sequence, we can uniquely identify the target of interest, even in the presence of multiple spatially-distributed instances.
	\end{itemize}
	
	\section{Training}
	\label{sec:training}
	
	
	
	\subsection{Datasets}
	
	The development of robust spatial intelligence in MLLMs is fundamentally constrained by existing datasets, which exhibit significant deficiencies in scale, task diversity, modality, and consistency. Current datasets lack the necessary volume to instill a generalized "spatial sense" and offer constrained task diversity, rarely incorporating structured scene reasoning. Furthermore, non-unified coordinate systems across disparate metadata sources hinder effective joint training. To bridge these gaps, we propose a large-scale, unified spatial intelligence training dataset.
	
	\begin{figure}[h]
		\centering
		\begin{subfigure}{0.38\textwidth}
			\centering
			\includegraphics[width=\linewidth]{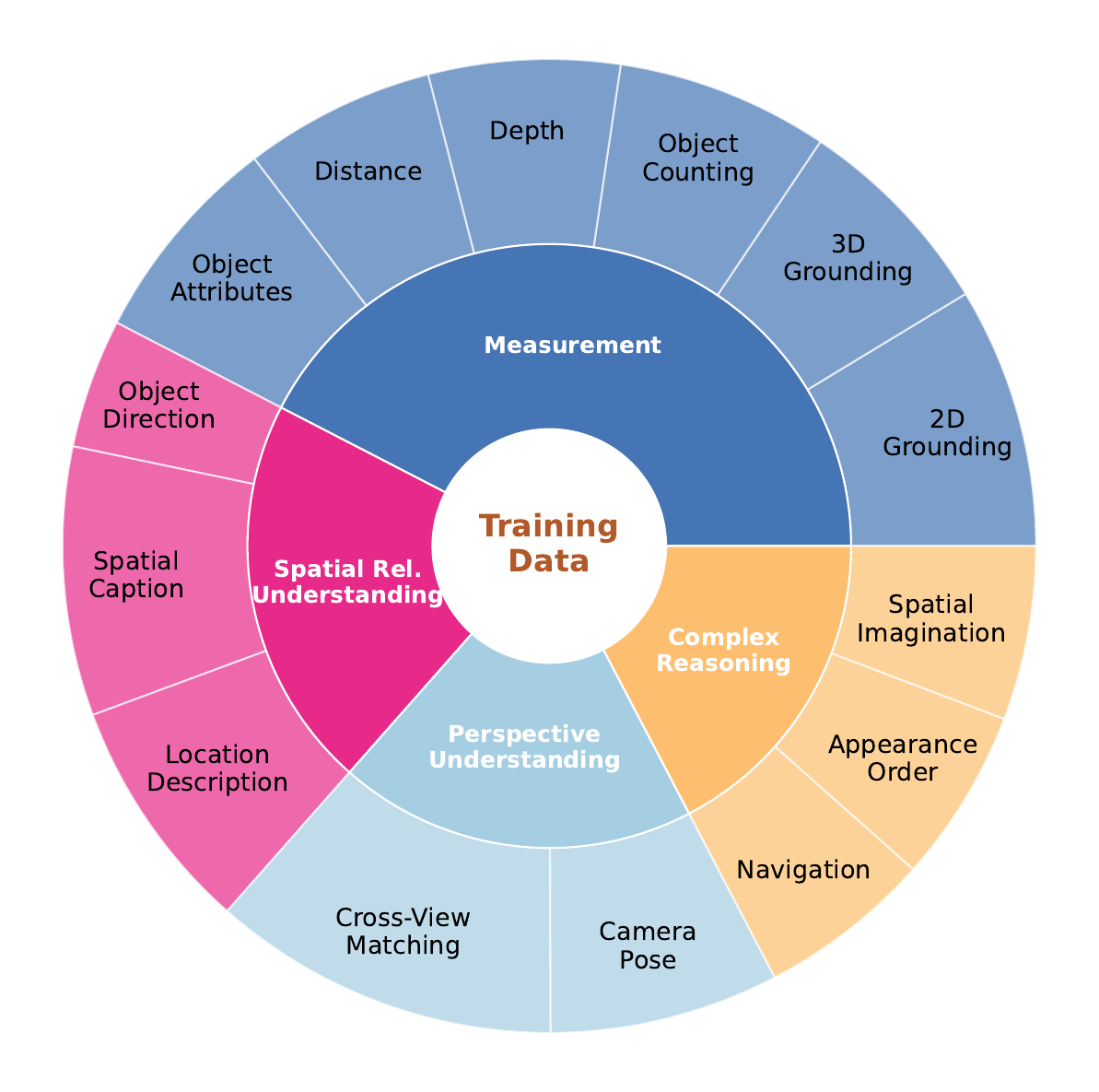}
		\end{subfigure}
		\hfill
		\begin{subfigure}{0.6\textwidth}
			\centering
			\includegraphics[width=\linewidth]{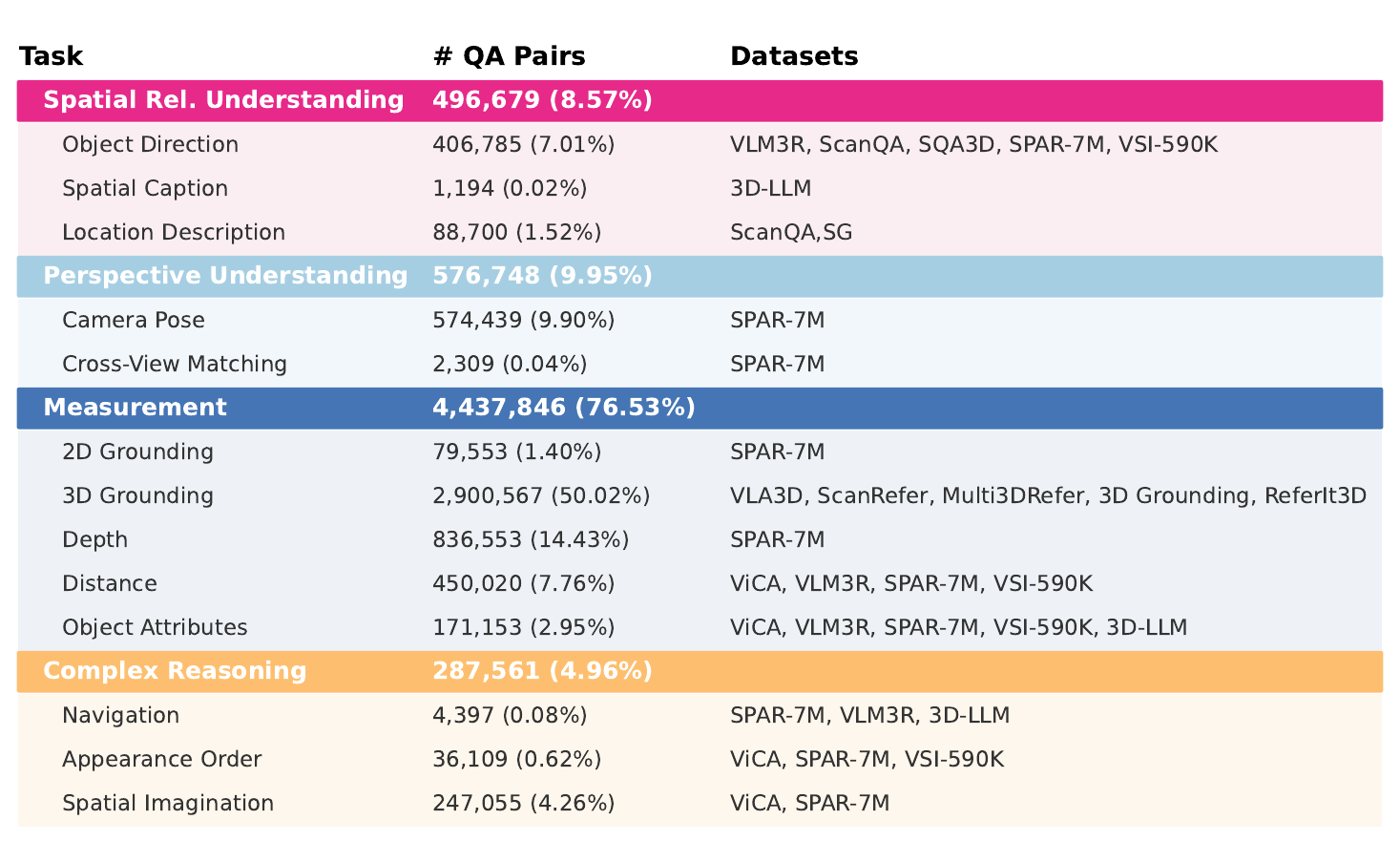}
		\end{subfigure}
		\caption{Task capability distribution of the datasets. Training samples are hierarchically organized into 4 primary categories and 14 fine-grained subtasks, encompassing a comprehensive spectrum of spatial reasoning capabilities.}
		\label{fig:data_chart}
	\end{figure}
	
	
	\noindent\textbf{Task Taxonomy.} As illustrated in Fig.~\ref{fig:data_chart}, our dataset encompasses a broad spectrum of spatial challenges organized into four primary categories. \textbf{Spatial Relationship Understanding:} Focuses on the relative positioning of objects, including directional relationships, spatial captioning, and precise location description. \textbf{Perspective Understanding}: Targeted at decyphering camera orientations, including camera pose estimation and cross-view matching. \textbf{Measurement:} Develops the model's ability to reason over quantitative metrics, such as 2D/3D grounding, depth estimation, absolute distance, and geometric object attributes. \textbf{Complex reasoning:} Cultivates high-level cognitive abilities, including navigation, appearance sequencing, and spatial imagination.
	
	
	
	\begin{figure}[htbp]
		\centering
		\begin{subfigure}[b]{0.98\textwidth}
			\centering
			\includegraphics[angle=270, width=0.95\linewidth]{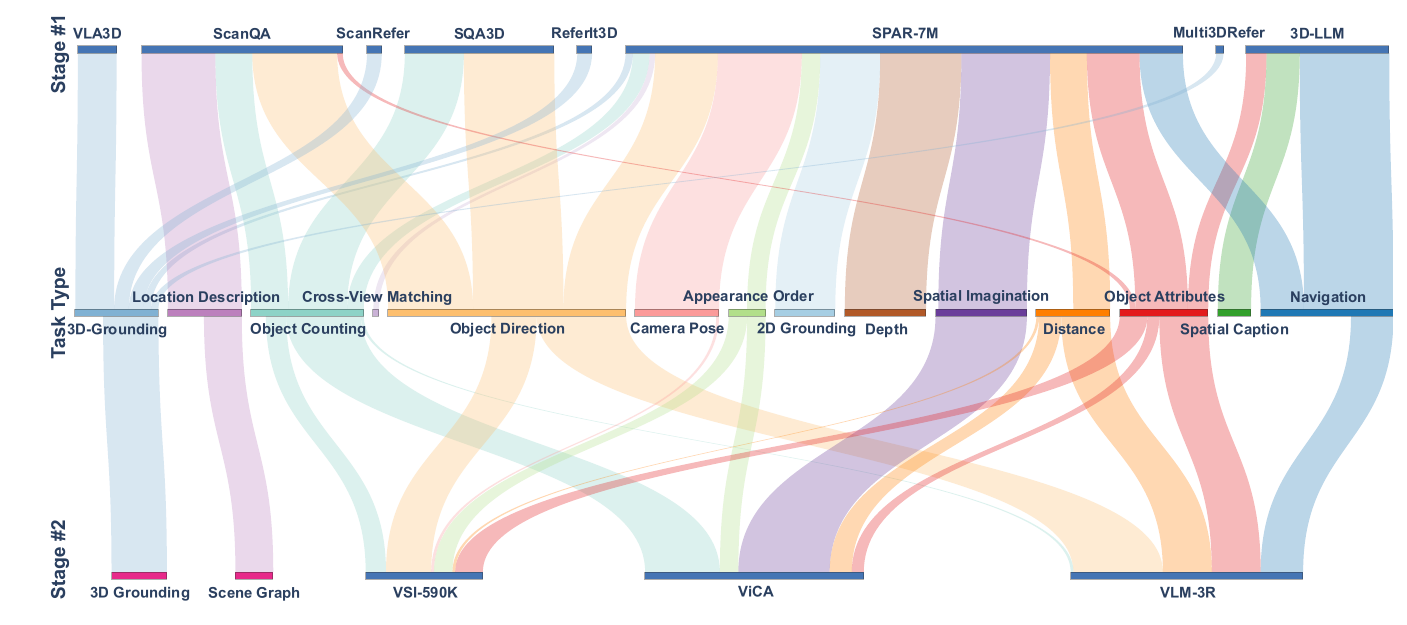}
		\end{subfigure}
		\begin{subfigure}[b]{0.98\textwidth}
			\centering
			\includegraphics[angle=0, width=0.95\linewidth]{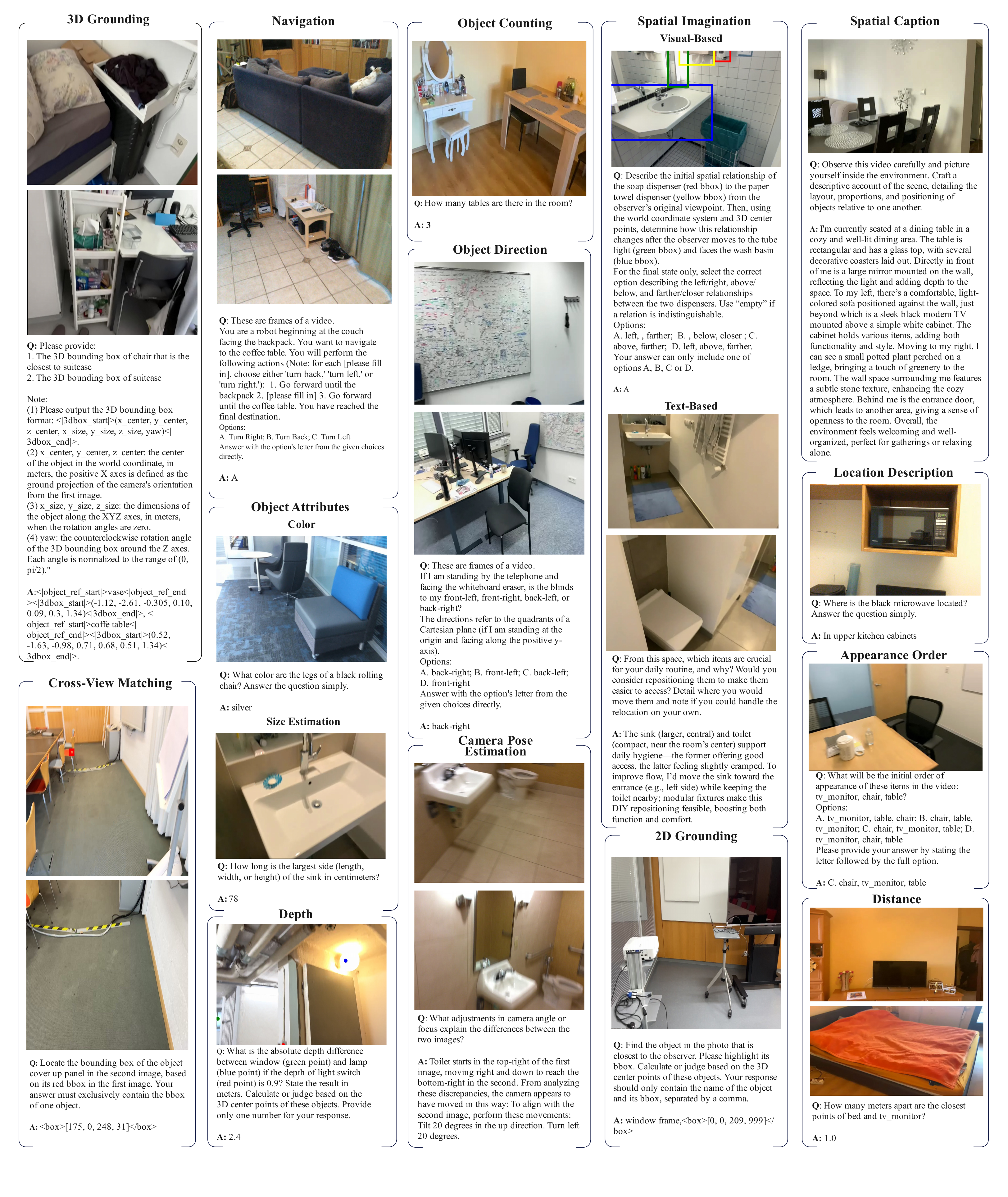}
		\end{subfigure}
		\vspace{-20pt}
		\caption{Taxonomy of training data used in the two training stages and representative QA pairs.}
		\label{fig:data_distribution_example}
	\end{figure}
	
	\subsubsection{Open Source Data Curation.} To bridge the gap between fundamental grounding and high-level spatial reasoning, we curate a diverse collection of open-source datasets. We incorporate a balanced subset of 3M QA pairs from SPAR-7M~\cite{zhang2025flatland} alongside general QA datasets (e.g., 3DLLM~\cite{hong20233dllm}, SQA3D~\cite{ma2022sqa3d}, ScanQA~\cite{azuma2022scanqa}) to broaden task diversity. Crucially, we unify the 3D global grounding coordinates from VLA3D~\cite{zhang2024vla3d}, ScanRefer~\cite{chen2020scanrefer}, ReferIt3D~\cite{achlioptas2020referit_3d}, and Multi3DRefer~\cite{zhang2023multi3drefer} into a canonical coordinate system. To address deficiencies in scene realism and spatio-temporal complexity, we also integrate high-quality reasoning data from ViCA~\cite{feng2025vica}, VLM3R~\cite{fan2025vlm3r}, and the real-world subset of VSI-590K~\cite{yang2025cambrians}. 
	
	A comprehensive description of our task definitions and the dataset normalization process is provided in the Appendix S1.
	
	\subsection{Training Strategy}
	\label{sec:train_strategy}
	
	We instantiate two distinct model variants: \textbf{\sysname-3D}, our comprehensive architecture (Fig.~\ref{fig:29}) that fuses 2D and 3D features, and \textbf{\sysname-2D}, a streamlined version utilizing only 2D visual inputs by omitting spatial features and embeddings. This dual-variant design ensures deployment flexibility while maintaining robust performance across disparate modality configurations. A visual comparison of these variants and their respective architectural components is provided in Appendix Fig. S2.
	
	We propose a two-stage training protocol to optimize efficiency and representation learning. In Stage 1, we train the \sysname-2D model exclusively with around 5.6M data samples. This strategy reduces the computational burden of large-scale training with 3D features while facilitating the learning of generalized representations; the 2D model operates on raw visual inputs, while the 3D model later integrates these with spatial features extracted via the VGGT backbone. In Stage 2, the \sysname-3D model is initialized with weights from the pre-trained 2D variant, while spatial-specific parameters are trained from scratch with around 917K data samples.
	
	To progressively cultivate the model's spatial intelligence, we employ a curriculum of increasing complexity. Fig.~\ref{fig:data_distribution_example} shows the data distribution across the two training stages. Stage 1 focuses on fundamental cognitive capabilities leveraging a diverse set of open-sourced datasets including SPAR-7M~\cite{zhang2025flatland}, 3DLLM~\cite{hong20233dllm}, SQA3D~\cite{ma2022sqa3d}, ScanQA~\cite{azuma2022scanqa}, VLA3D~\cite{zhang2024vla3d}, ScanRefer~\cite{chen2020scanrefer}, ReferIt3D~\cite{achlioptas2020referit3d}, and Multi3DRefer~\cite{zhang2023multi3drefer}. Stage 2 targets high-level structured scene reasoning, specifically scene graph generation and global-scale 3D grounding. This stage incorporates our custom generation pipeline alongside open-source datasets like ViCA~\cite{feng2025vica}, VLM3R~\cite{fan2025vlm3r}, and VSI-590K~\cite{yang2025cambrians}. We exclusively utilize the training splits of these open-source datasets and ensure that all video sequences are strictly isolated from the VSI-Bench evaluation suite to prevent data contamination. Detailed training hyperparameters and strategies are provided in the Appendix S3.

	\section{Experiments}
	\label{sec:exp}
	
	\subsection{Implementation Details}
	We adopt openPangu-VL-7B~\cite{openpanguvl7b} as our base model. The training was conducted using 128 Ascend 910B3 NPUs. For all video inputs, we utilize a uniform temporal sampling strategy to extract 32 frames. The training protocol follows a two-stage schedule, consisting of 1 epoch for the first stage and 3 epochs for the second stage.
	
	\subsubsection{Baselines.}
	To rigorously assess the performance of \sysname, we benchmark it against three distinct architectural categories: (1) \textbf{Proprietary models}, including state-of-the-art (SOTA) vision-language models such as GPT-5~\cite{singh2025gpt5}, Gemini-3 Pro~\cite{gemini2025gemini3pro}, etc; (2) \textbf{Open-source general MLLMs}, featuring high-capacity generalists such as InternVL3.5~\cite{wang2025internvl35}, Qwen3-VL~\cite{Qwen3-VL}, long-context architectures~\cite{zhang2024longva,chen2024longvila}, etc; and (3) \textbf{Specialized spatial MLLMs}, Domain-specific models tailored for spatial intelligence, featured by MindCube~\cite{yin2025spatial}, GS-Reasoner~\cite{chen2026reasoning}, and VLM-3R-7B~\cite{fan2025vlm3r}, etc. This diverse selection ensures a robust comparison against both general-purpose reasoning and domain-specific spatial intelligence.
	
	\subsubsection{Benchmarks.}
	In addition to VSI-Bench~\cite{yang2025thinking}, we evaluate our model across five complementary benchmarks to capture the full spectrum of spatial intelligence. These include \textbf{VSI-Bench$^{\textbf{Debiased}}$}~\cite{yang2025thinking}, which targets model robustness through challenging spatial queries; \textbf{MindCube}~\cite{yin2025spatial} and \textbf{ViewSpatial}~\cite{li2025viewspatialbench}, which assess cross-view spatial reasoning and latent 3D structure inference across diverse indoor and outdoor settings; \textbf{SpaCE-10}~\cite{gong2025space10}, focusing on spatial configuration consistency within image sequences; and \textbf{VSTI-Bench}~\cite{fan2025vlm3r}, which evaluates fine-grained spatio-temporal camera-object relationships. This diverse suite ensures a rigorous validation of perception, reasoning, and temporal consistency. Evaluation on MindCube, ViewSpatial, and SpaCE-10 is limited to the \sysname-2D variant, as these datasets do not provide temporal video inputs.
	Comprehensive list of evaluated models are detailed in the Appendix S4.
	
	
	\subsection{Main Results}
	
	\begin{table}[t]
		\centering
		\caption{Spatial and spatiotemporal intelligence performance on key benchmarks.}
		\label{tab:exp_overall}
		\resizebox{\linewidth}{!}{
			\begin{tabular}{l l c c c c c c}
				\toprule
				\textbf{Category} & \textbf{Model} & \makecell[c]{VSI-Bench~\cite{yang2025thinking}} & \makecell[c]{VSI-Bench$^{\textbf{Debiased}}$~\cite{yang2025cambrians}} & \makecell[c]{MindCube~\cite{yin2025spatial}} & \makecell[c]{ViewSpatial~\cite{li2025viewspatialbench}}  & \makecell[c]{SpaCE-10~\cite{gong2025space10}} & \makecell[c]{VSTI-Bench~\cite{fan2025vlm3r}} \\
				\midrule
				\multirow{3}{*}{Baseline} & Human & 79.2† & - & 94.5† & - & 91.2† & 77.0† \\
				& Random Choice & 34.0†  & - & 33.0† & 26.3† & - & - \\
				& Chance Level (Frequency) & -& - & - & -  & - & 22.4† \\
				\midrule
				\multirow{5}{*}{\makecell[l]{Proprietary\\models}} & Seed-1.6~\cite{guo2025seed15vl} & 49.9  & - & 48.7 & 43.8 &  -& - \\
				& Grok-4~\cite{grok4announcement} & 47.9  & - & 63.5 & 43.2 & - & - \\
				& GPT-5~\cite{singh2025gpt5} & 55.0 & - & 56.3 & 45.5  & 53.4 & - \\
				& Claude-3.7-Sonnet (cla)~\cite{anthropic2024claude3} & 47.0 & - & - & - & 46.2 & - \\
				& Gemini-3 Pro~\cite{gemini2025gemini3pro} & 52.5 & - & \underline{70.8} & 50.3 & - & - \\
				\midrule
				\multirow{6}{*}{\makecell[l]{Open-Sourced\\general models}} & Bagel-7B-MoT & 31.4 & - & 34.7 & 41.3 & - & -\\
				& Qwen3-VL-235B-A22B-Instruct~\cite{Qwen3-VL} & 62.7† &  - &  - & - & 47.9 & -\\
				& InternVL3.5-241B-A28B~\cite{wang2025internvl35} & 69.5 & - &  - &  -  & \underline{55.0} & -\\
				& LLaVA-OneVision-7B~\cite{li2024llavaonevision} & 32.4  & 28.5  & - & - & - & 41.7 \\
				& LLaVA-Video-7B~\cite{zhang2024video} & 35.6 & 30.7 & - & - & -& - \\
				& SmolVLM2-2.2B~\cite{marafioti2025smolvlm} & 27.0 & 22.3  & - & -  & - & -\\
				\midrule
				\multirow{12}{*}{\makecell[l]{Open-Sourced\\spatial models}} & MindCube-3B-RawQA-SFT~\cite{yin2025spatial} & 17.2 & - & 51.7 & 24.1  & - & -\\
				& SpatialLadder-3B~\cite{li2025spatialladder} & 50.8† & - & 27.4 & 44.2† & -& - \\
				& Spatial-MLLM-4B~\cite{wu2025spatialmllm} & 48.4† & - & 26.1 & 34.6  & -& - \\
				& SpaceR-7B~\cite{ouyang2025spacer} & 45.6† & - & 27.4  & 35.8 & -  & -\\
				& ViLaSR-7B~\cite{wu2025reinforcing} & 45.4† & - & 30.2†  & 35.7 & -& - \\
				& VLM-3R-7B~\cite{fan2025vlm3r} & 60.9 & - & 40.0 & 40.5 & - & \textbf{58.8} \\
				& GS-Reasoner~\cite{chen2026reasoning} & 64.7 & - & - & - & - & - \\
				& VST-7B-SFT~\cite{yang2025visual} & 61.2† & - & 32.0†  & 50.5 & -& - \\
				& Cambrian-S-7B~\cite{yang2025cambrians} & 67.5† & 56.3†  & 39.6 & 40.9 & - & -\\
				& SenseNova-SI (InternVL3-8B)~\cite{cai2025scaling} & 68.7 & 62.8 & \textbf{85.6} & \underline{54.6} & - & -\\
				& \textbf{\sysname-2D (Ours)} & \underline{71.9} & \underline{66.6} & 63.5 & \textbf{59.7}  & \textbf{65.7} & 41.8 \\
				&\textbf{\sysname-3D (Ours)} & \textbf{73.9} & \textbf{69.9} & - & -  & - & \underline{44.8} \\
				\bottomrule
			\end{tabular}
		}
		\begin{tablenotes}[footnotesize]
			\centering
			\parbox{\linewidth}{\item[-]- No publicly available data; † Data cited from the original model paper. Other sources: \cite{cai2025scaling, yang2025cambrians, fan2025vlm3r}.}
		\end{tablenotes}
	\end{table}
	
	\begin{table}[t]
		\centering
		\caption{Comparison with state-of-the-art MLLMs on VSI-Bench. \sysname~achieves the best performance.}
		\label{tab:exp_VSI-Bench}
		\begin{threeparttable}
			\resizebox{\linewidth}{!}{
				\begin{tabular}{l l|c c c c c c c c|c}
					\toprule
					\textbf{Category} & \textbf{Model} & \makecell[c]{\rotatebox{30}{Rel. Dir.}} & \makecell[c]{\rotatebox{30}{Rel. Dist.}} & \makecell[c]{\rotatebox{30}{Appr. Order}} & \makecell[c]{\rotatebox{30}{Route Plan}} & \makecell[c]{\rotatebox{30}{Obj. Size}}  & \makecell[c]{\rotatebox{30}{Obj. Count}} & \makecell[c]{\rotatebox{30}{Abs. Dist.}} & \makecell[c]{\rotatebox{30}{Roome Size}} & \textbf{\rotatebox{0}{Overall}} \\
					\midrule
					\multirow{1}{*}{\textbf{Baseline}}
					& Human~\cite{yang2025thinking} & 95.8 & 94.7 & 100 & 95.8 & 60.4 & 94.3 & 47.0 & 45.9 & 79.2 \\
					\midrule
					\multirow{1}{*}{\makecell[l]{\textbf{Proprietary Models}}}
					& GPT-5-2025-08-07~\cite{singh2025gpt5} & 48.6 & 63.7 & 68.9 & \cellcolor{myred!30}50.2 & 73.3 & 53.5 & 34.4 & 47.5 & 55.0 \\
					\midrule
					\multirow{5}{*}{\makecell[l]{\textbf{Open-Sourced}\\\textbf{General Models}}}
					& LLaVA-OneVision-7B~\cite{li2024llavaonevision} & 35.2 & 42.5 & 24.4 & 29.4 & 47.4 & 47.7 & 20.2 & 12.3 & 32.4 \\
					& LLaVA-Video-7B~\cite{zhang2024video}& 42.4 & 43.5 & 30.6 & 34.0 & 47.8 & 48.5 & 14.0 & 24.2 & 35.6 \\
					& Qwen3-VL-8B-Instruct~\cite{Qwen3-VL} & 50.9 & 58.0 & 66.3 & 35.0 & \cellcolor{myred!10}76.3 & 53.3 & 47.0 & 61.9 & 57.9 \\
					& InternVL3-8B~\cite{wang2025internvl35} & 39.3 & 48.0 & 31.3 & 26.2 & 43.6 & 66.0 & 34.8 & 47.5 & 42.1 \\
					\midrule
					\multirow{8}{*}{\makecell[l]{\textbf{Open-Sourced}\\\textbf{Spatial Models}}}
					& SpaceR-7B~\cite{ouyang2025spacer} & 46.1 & 41.9 & 54.8 & 29.3 & 53.5 & 44.5 & 24.7 & 37.3 & 41.5 \\
					& ViLaSR-7B~\cite{wu2025reinforcing} & 46.5 & 45.0 & 53.2 & 29.9 & 61.4 & 58.1 & 33.8 & 28.8 & 44.6 \\
					& VLM-3R-7B~\cite{fan2025vlm3r} & 80.5 & 65.4 & 40.1 & 45.4 & 69.2 & 70.2 & 49.4 & 67.1 & 60.9 \\
					& ViCA-7B~\cite{feng2025vica} & 42.6 & 58.5 & 68.8 & 34.5 & \cellcolor{myred!60}\textbf{79.2} & 68.8 & 57.0 & \cellcolor{myred!10}75.1 & 60.6 \\
					& VST-7B~\cite{yang2025visual} & 55.6 & 60.0 & 69.2 & 44.3 & 75.5 & 71.6 & 43.8 & 69.2 & 61.2 \\
					&  GS-Reasoner (pred dep.)~\cite{chen2026reasoning} & \cellcolor{myred!30}88.9 & 65.4 & 52.3 & 44.3 & 70.0 & 69.1 & 61.9 & 65.7 & 64.7 \\
					& Cambrian-S-7B~\cite{yang2025cambrians} & 76.2 & \cellcolor{myred!10}71.1 & \cellcolor{myred!10}80.1 & 41.8 & 74.9 & \cellcolor{myred!30}73.2 & 50.5 & 72.2 & 67.5 \\
					& SenseNova-SI \textsmaller{InternVL3-8B}~\cite{cai2025scaling} & 80.7 & \cellcolor{myred!60}\textbf{76.3} & 48.4 & \cellcolor{myred!60}\textbf{69.5} & 72.7 & \cellcolor{myred!60}\textbf{76.7} & \cellcolor{myred!60}\textbf{72.0} & 53.5 & \cellcolor{myred!10}68.7 \\
					& \textbf{\sysname-2D~(ours)} & \cellcolor{myred!10}\textbf{87.9} & 69.2 & \cellcolor{myred!30}\textbf{81.9} & \cellcolor{myred!10}48.5 & \cellcolor{myred!30}76.5  & \cellcolor{myred!10}71.8 & \cellcolor{myred!10}63.4 & \cellcolor{myred!30}\textbf{76.5} & \cellcolor{myred!30}\textbf{71.9} \\
					& \textbf{\sysname-3D~(ours)} & \cellcolor{myred!60}\textbf{93.4} & \cellcolor{myred!30}71.3 & \cellcolor{myred!60}\textbf{85.0} & \cellcolor{myred!10}48.5 & 76.0  & 70.5 & \cellcolor{myred!30}67.1 & \cellcolor{myred!60}\textbf{79.5} & \cellcolor{myred!60}\textbf{73.9} \\
					\bottomrule
				\end{tabular}
			}
		\end{threeparttable}
	\end{table}

	\subsubsection{Performance on Spatial Intelligence Benchmarks.}
	As summarized in Tab.~\ref{tab:exp_overall}, \sysname~achieves SOTA performance across the majority of evaluated benchmarks. On VSI-Bench, the \sysname-3D variant attains a score of 73.9, surpassing the previous SOTA, InternVL3.5-241B \cite{wang2025internvl35}, by a margin of 4.4 points. Remarkably, even without 3D spatial input features, our \sysname-2D variant outperforms InternVL3.5-241B by 2.4 points. This advantage is further amplified on SpaCE-10 \cite{gong2025space10}, where the 2D model alone exceeds InternVL3.5-241B by 10.7 points. These results, achieved with only 7 billion parameters, demonstrate that targeted training on large-scale spatial reasoning tasks allows compact models to significantly outperform their much larger, general-purpose counterparts.
	
	On VSI-Bench$^{\textbf{Debiased}}$ \cite{yang2025cambrians}—a subset specifically designed to eliminate questions answerable via linguistic priors—\sysname-3D surpasses Cambrian-S \cite{yang2025cambrians} by 13 points, while \sysname-2D maintains an 11-point lead. This underscores the robustness of our visually grounded reasoning framework. Furthermore, despite lacking specialized spatio-temporal architectural components, \sysname~achieves competitive results on VSTI-Bench \cite{fan2025vlm3r}. Finally, \sysname-2D attains a score of 61.6 on ViewSpatial, outperforming SenseNova-SI \cite{cai2025scaling} by 5 points, and secures a top-3 ranking on MindCube \cite{yin2025spatial} despite not being exposed to its training split. These results validate the model’s ability to perform multi-image spatial reasoning under constrained viewpoints, highlighting the efficacy of our LocalCogMap formulation in capturing complex layouts from 2D inputs without auxiliary geometric signals.
	
	\subsubsection{In-Depth Analysis on VSI-Bench.}
	VSI-Bench serves as a canonical evaluation suite for indoor spatial reasoning via video sequences. We compare \sysname~against leading proprietary and open-source models, restricting the latter to those with comparable parameter counts (7B–8B). As shown in Tab.~\ref{tab:exp_VSI-Bench}, \sysname-2D achieves a score of 71.9, outperforming SenseNova-SI (InternVL2-8B) \cite{cai2025scaling} by 3.2 points. The full \sysname-3D variant further elevates this score to 73.9, establishing a 5.2-point lead over SenseNova-SI.
	We attribute these gains to two primary design choices: (1) our 3D feature fusion branch, which injects multi-view consistent geometric cues, and (2) structured reasoning objectives, specifically scene graph generation and 3D global grounding. By supervising the model to explicitly reconstruct scene layouts, we foster a "spatial-first" reasoning paradigm where the mental modeling of 3D structures precedes high-level semantic inference.
	
	Notably, \sysname~surpasses human-level performance in metric-estimation tasks such as Object Size, Room Size, and Absolute Distance. This phenomenon stems from a cognitive divergence: while humans rely on qualitative spatial heuristics and often struggle with precise metric estimation from visual memory, \sysname~successfully internalizes these quantitative spatial distributions by leveraging our large-scale structured dataset.
	
	

	\subsubsection{Performance in 3D Grounding.} During the first training stage, we utilize large-scale 3D global grounding as an auxiliary task to bolster the model’s spatial reasoning foundations. To evaluate the efficacy of this pre-training, we build a 7-DoF 3D global grounding test set consisting of 10,000 QA pairs derived from the ScanNet \cite{dai2017scannet} test split.
	We compare our model against Qwen3-VL \cite{Qwen3-VL} as a primary baseline. As illustrated in Fig.~\ref{fig:3d_grounding}, our model significantly outperforms Qwen3-VL; our prediction errors are tightly clustered within the [0, 0.7] range, whereas Qwen3-VL’s errors predominantly fluctuate between [1.0, 1.7]. This substantial reduction in error indicates that our specialized pre-training effectively calibrates the model's understanding of absolute 3D spatial coordinates.
	
	\subsubsection{Performance in LocalCogMap Prediction.} 
\begin{figure}[t] 
		\centering
		\begin{subfigure}{0.48\linewidth}
			\centering
			\includegraphics[width=\linewidth]{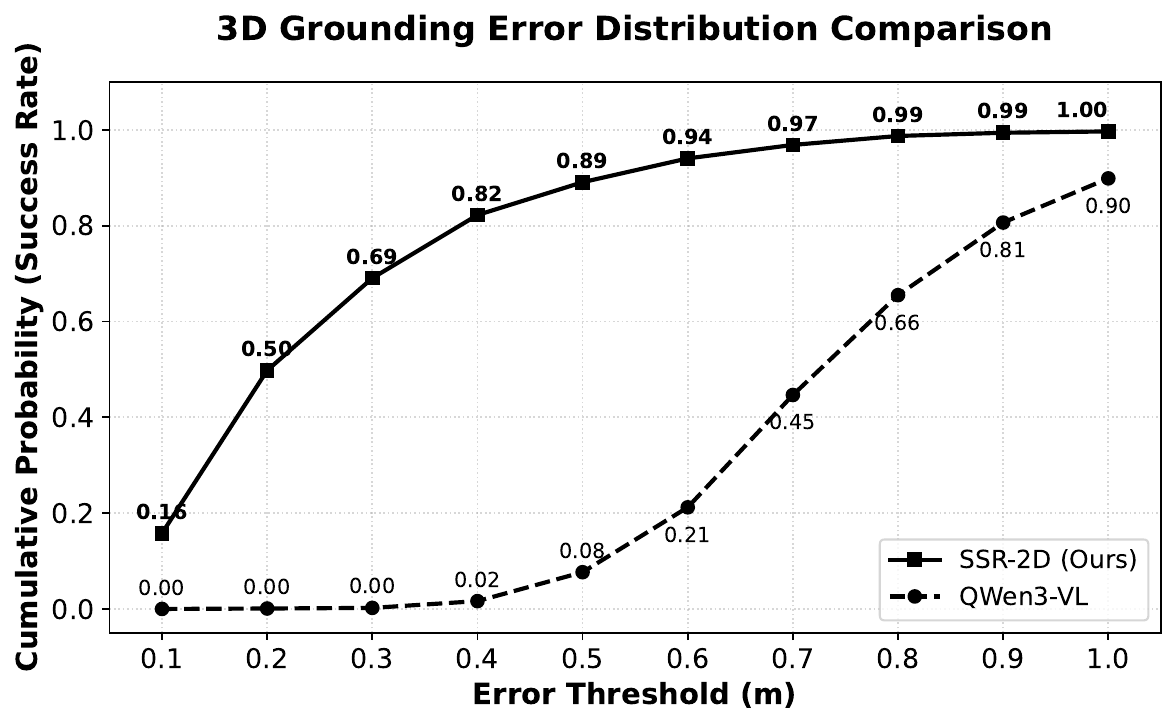}
			\caption{3D global grounding comparison.}
			\label{fig:3d_grounding}
		\end{subfigure}
		\hfill 
		\begin{subfigure}{0.48\linewidth}
			\centering
			\includegraphics[width=\linewidth]{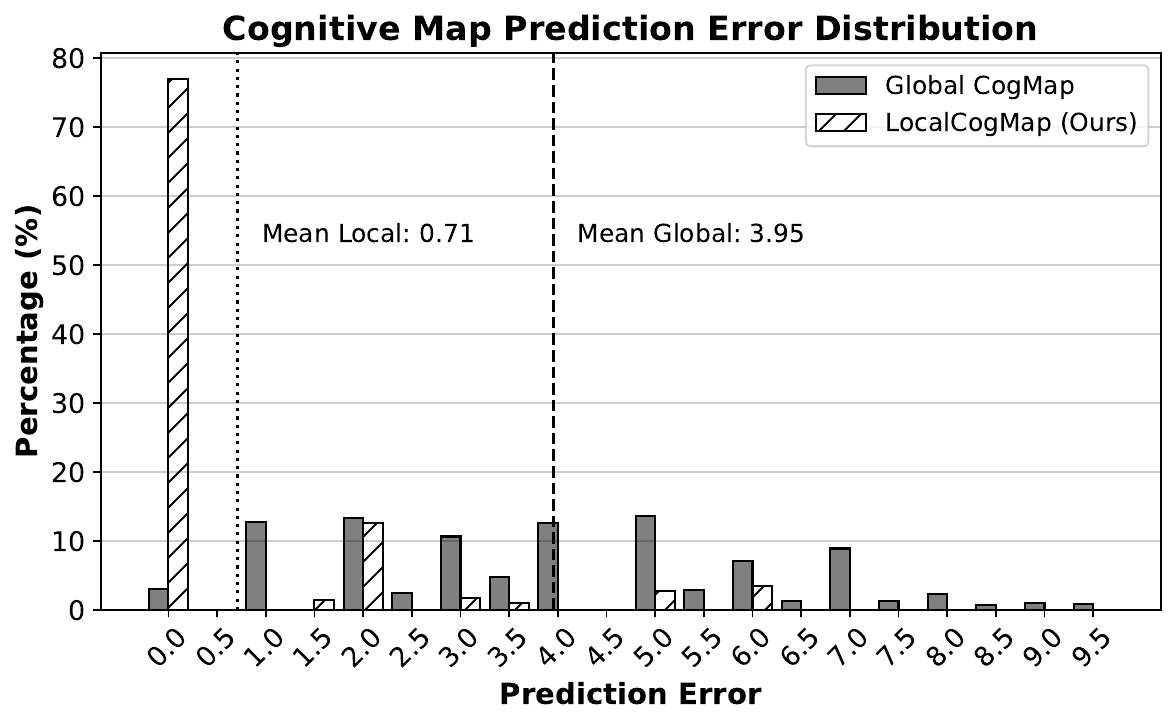}
			\caption{CogMap error distribution.}
			\label{fig:cogmap_error}
		\end{subfigure}
		
		\caption{Experimental results. (Left) The 3D global grounding comparison results between \sysname-2D and Qwen3-VL. (Right) The error histogram comparison between LocalCogMap and global CogMap introduced in VSI-Bench.}
		\label{fig:combined_results}
	\end{figure}

In VSI-Bench, a global cognitive map was introduced to project scene layouts onto a $10 \times 10$ grid. However, we observe that the restricted viewpoints typical of video sequences make it exceedingly difficult for models to maintain accurate global distributions. In this experiment, we evaluate the spatial localization error of the global cognitive map against our proposed LocalCogMap. As shown in Fig.~\ref{fig:cogmap_error}, the LocalCogMap achieves a mean prediction error of only 0.71 units, significantly lower than the global competitor. This performance disparity confirms that predicting object layouts within a localized coordinate system is a far more tractable objective for the model. By shifting from a global to a local framework, the task better aligns with the incremental nature of visual perception in video data, where spatial context is built progressively rather than captured in a single, all-encompassing view.
	
	\subsection{Ablation Study}
	
	\begin{table}[t]
		\centering
		\caption{Ablation studies of \sysname~on VSI-Bench concerning model components and training data. The gray row (\cmark) represents our default/best configuration used across experiments.}
		\label{tab:block_merged}
		\setlength{\tabcolsep}{4pt} 
		\resizebox{\linewidth}{!}{
			\begin{tabular}{l cccccccc c}
				\toprule
				\textbf{Method/Config} & \makecell[c]{\rotatebox{30}{Rel. Dir.}} & \makecell[c]{\rotatebox{30}{Rel. Dist.}} & \makecell[c]{\rotatebox{30}{Appr. Order}} & \makecell[c]{\rotatebox{30}{Route Plan}} & \makecell[c]{\rotatebox{30}{Obj. Size}}  & \makecell[c]{\rotatebox{30}{Obj. Count}} & \makecell[c]{\rotatebox{30}{Abs. Dist.}} & \makecell[c]{\rotatebox{30}{Roome Size}} & \textbf{\rotatebox{0}{Overall}} \\
				\midrule
				\rowcolor[gray]{0.85} \sysname-2D (Default) & 87.9 & 69.2 & 81.9 & 48.5 & 76.5 & 71.8 & 63.4 & 76.5 & \textbf{71.9} \\
				
				\midrule
				\multicolumn{10}{l}{\textit{(a) Token Insertion Method (\sysname-3D)}} \\
				Sequential & 88.5 & 69.2 & 80.7 & 49.5 & 76.3 & 68.7 & 65.2 & 78.9 & 72.1 \\
				Interleaved (\cmark) & 93.4 & 71.3 & 85.0 & 48.5 & 76.0 & 70.5 & 67.1 & 79.5 & \textbf{73.9} \\
				
				\addlinespace[5pt]
				\multicolumn{10}{l}{\textit{(b) Training Phases (\sysname-2D)}} \\
				w/ Stage 1 (\cmark) & 87.9 & 69.2 & 81.9 & 48.5 & 76.5 & 71.8 & 63.4 & 76.5 & \textbf{71.9} \\
				w/o Stage 1 & 83.8 & 69.0 & 69.1 & 43.3 & 77.6 & 68.5 & 60.0 & 79.1 & 68.8 \\
				
				\addlinespace[5pt]
				\multicolumn{10}{l}{\textit{(c) Training Data Composition (\sysname-2D)}} \\
				Base Data & 86.4 & 69.4 & 72.5 & 40.2 & 76.6 & 73.4 & 62.5 & 75.9 & 69.6 \\
				+ Grounding (GRD) & 85.4 & 67.9 & 74.6 & 47.9 & 76.7 & 72.7 & 62.8 & 75.8 & 70.5 \\
				+ Scene Graph (SG) & 87.1 & 72.5 & 80.9 & 40.2 & 77.4 & 71.1 & 63.8 & 77.0 & 71.2 \\
				+ SG + GRD (\cmark) & 87.9 & 69.2 & 81.9 & 48.5 & 76.5 & 71.8 & 63.4 & 76.5 & \textbf{71.9} \\
				\bottomrule
			\end{tabular}
		}
	\end{table}
	
	\subsubsection{Ablation Study on Token Insertion Methods.} To evaluate the effectiveness of our proposed interleaved token insertion strategy, we conduct an ablation study using the same \sysname-3D architecture under two token insertion schemes: (1) \textbf{sequential insertion}, where visual and spatial tokens are concatenated in separate blocks, and (2) \textbf{interleaved insertion}, our proposed method that alternates visual and spatial tokens throughout the sequence. As shown in Tab.~\ref{tab:block_merged}, the choice of token insertion plays a critical role in aligning visual and spatial features. Specifically, transitioning from sequential to interleaved insertion elevates the 3D model’s performance from 72.1 to 73.9. This shift not only amplifies its competitive edge over the 2D baseline but also underscores the necessity of fine-grained cross-modal interaction in fortifying the model’s local spatial reasoning capabilities.
	
	\subsubsection{Ablation Study on Training Phase.} In our training strategy, we split the training pipeline into two training phases. The training stage 1 is designed to equipped the model with basic reasoning and grounding abilities, setting up a strong base model to be ready to learn complex tasks in the following stages. In the second stage, we train the model with complex spatial reasoning tasks and strustured reasoning tasks (scene graph generation and 3D global grounding), which directly enhances the capbilities tightly related to that in VSI-Bench and the other spatial reasoning benchmarks. In this ablation study, we investigate whether the training stage 1 is neccessary. In another word, is it possible for the model to diretly learn complex tasks from scratch. From Tab.~\ref{tab:block_merged}, we can learn that without training stage 1, the performance decreases sharply from 71.9 to 68.8, meaning that a step-by-step training can easily generalize the model capability from basic to complex manner. In future work, we envision that more stages of training play a more important role in building spatial intelligence foundation model.
	
	\subsubsection{Ablation Study on Data Composition.} In contrast to contemporary spatial intelligence models such as ViCA~\cite{feng2025vica} and VLM3R~\cite{fan2025vlm3r}, which focus primarily on the eight core tasks defined in VSI-Bench, our approach enriches the training stage 2 phase with structured auxiliary tasks, including scene graph generation and 3D global grounding. We hypothesize that cultivating spatial mental modeling—by training the model to generate sparse, symbolic abstractions of a scene—is a critical prerequisite for advanced spatial reasoning. To quantify the contribution of these structured tasks, we conducted an ablation study using the same training configuration in stage 1 as our primary experiments while excluding scene graph and 3D global grounding data from the second stage. As shown in Tab.~\ref{tab:block_merged}, the omission of these two tasks results in a significant performance drop on VSI-Bench, from 71.9 to 69.6. Notably, the Appearance Order and Route Planning tasks experienced the most substantial degradation. This suggests that forcing the model to construct structured representations of its environment directly enhances its ability to track object occurrences and compute viable navigation paths, confirming that symbolic spatial understanding serves as a robust scaffold for complex downstream reasoning.
	
	\subsection{Data Scaling}
\begin{figure}[t]
		\centering
		\includegraphics[width=0.4\linewidth]{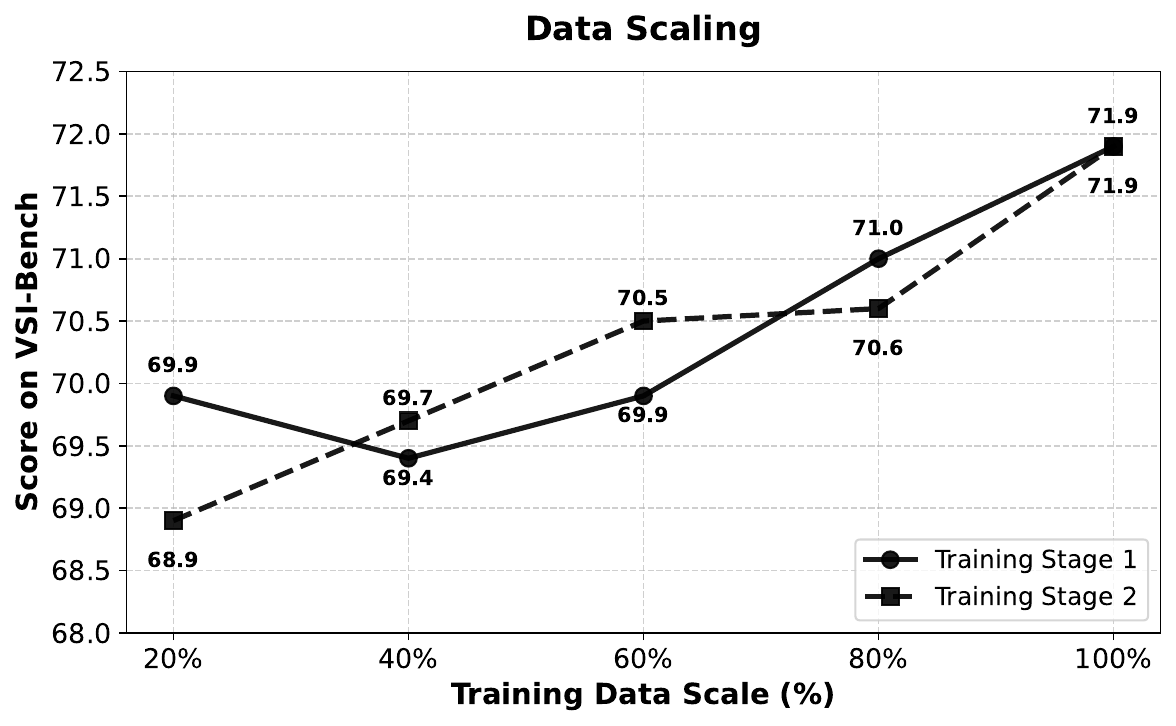}
		\caption{The figure shows the accuracy of \sysname~with the increasing of training data used in the two stages respecitvely. Overall, we can see that the accuracy of \sysname~increases with the increasing of training data.}
		\label{fig:exp_scaling}
	\end{figure}

	The concept of Scaling Laws is a cornerstone of modern LLM research, yet its applicability to the domain of spatial intelligence—particularly complex spatial reasoning—remains an open question. In this section, we conduct a data ablation study to empirically validate the existence of scaling phenomena in our framework. Specifically, we systematically increase the volume of training data across both of the training stages, scaling from 20\% to 100\% of the total dataset. As illustrated in Fig.~\ref{fig:exp_scaling}, model performance on VSI-Bench exhibits a steady, monotonic increase in correlation with data volume. This trend provides strong empirical evidence that scaling laws indeed govern the development of spatial intelligence, suggesting that further data expansion could continue to yield significant gains in reasoning proficiency.
	
	\section{Conclusion}
	\label{sec:conclusion}
	In this work, we presented \sysname, a specialized 7B-parameter spatial intelligence model designed to surmount the limitations of general-purpose MLLMs in complex geometric reasoning. By introducing a dual-branch 3D-aware architecture, we provide a lightweight yet effective multi-modal alignment paradigm. Furthermore, our structured scene reasoning paradigm, anchored by the LocalCogMap formulation, empowers the model to generate fine-grained "mental scene graphs" that serve as a robust cognitive foundation for complex tasks.
	While extensive evaluations confirm that \sysname~achieves leading results across competitive benchmarks—outperforming general-purpose models nearly 35 times its size—certain \textbf{limitations} remain. Specifically, the model's 3D awareness is constrained by its pre-training on 2D features, and the LocalCogMap is currently formulated in a bird's-eye-view perspective, which restricts its 3D representation capabilities. We expect future research to address these limitations by exploring more comprehensive 3D feature integration and volumetric scene representations.

	
	
	
	%
	%
	\bibliographystyle{splncs04}
	\bibliography{main.bib}
\end{document}

%% file: macro.tex
\newcommand{\sysname}{SSR}
\definecolor{myred}{RGB}{199,0,11}
\definecolor{mygrey}{RGB}{10,10,10}

\newcommand{\cmark}{\ding{51}} 